\documentclass{article}

\usepackage[final]{neurips_bdl2019}

\usepackage[utf8]{inputenc} 
\usepackage[T1]{fontenc}    
\usepackage{hyperref}       
\usepackage{url}            
\usepackage{booktabs}       
\usepackage{amsfonts}       
\usepackage{nicefrac}       
\usepackage{microtype}      

\usepackage{amsmath}        
\usepackage{graphicx}
\usepackage{bm}
\usepackage{subcaption}
\usepackage{color}

\usepackage{tikz}

\title{Bayesian Linear Regression on Deep Representations}

\author{%
  John Moberg\thanks{Correspondence to \texttt{john@moberg.ai}.}, \;Lennart Svensson, \;Juliano Pinto, \;Henk Wymeersch\\
  Chalmers University of Technology \\
}

\begin{document}

\maketitle

\begin{abstract}
  A simple approach to obtaining uncertainty-aware neural networks for regression is to do Bayesian linear regression (BLR) on the representation from the last hidden layer. Recent work \citep{Riquelme2018DeepSampling,Azizzadenesheli2018EfficientQ-networks} indicates that the method is promising, though it has been limited to homoscedastic noise. In this paper, we propose a novel variation that enables the method to flexibly model heteroscedastic noise. The method is  benchmarked against two prominent alternative methods on a set of standard datasets, and finally evaluated as an uncertainty-aware model in model-based reinforcement learning. Our experiments indicate that the method is competitive with standard ensembling, and ensembles of BLR outperforms the methods we compared to.
\end{abstract}

\section{Introduction}

Obtaining reliable uncertainty estimates from deep neural networks (NNs) is an active field of research with an abundance of proposed methods, e.g., variational inference, stochastic gradient MCMC, and ensembling. Two especially prominent methods are Monte-Carlo dropout (MC-dropout, \cite{Gal2016DropoutLearning}) and Deep Ensembles \citep{Lakshminarayanan2017SimpleEnsembles}, but MC-dropout appears less reliable and Deep Ensembles requires training several copies of the same model, which can be expensive. While many methods have been proposed, there is not yet a clear winner.

Recently, the simple idea of doing Bayesian linear regression (BLR) on the representation in the last hidden layer has caught some attention, e.g., for Bayesian optimization \citep{Snoek2015ScalableNetworks}, Thompson sampling in contextual bandits \citep{Riquelme2018DeepSampling}, and exploration in model-free reinforcement learning (RL) \citep{Azizzadenesheli2018EfficientQ-networks}. While not as general as some other methods, it is a simple and fast way to obtain uncertainty estimates from NNs with a linear and fully-connected output layer. Despite its application in several recent papers, its performance has not yet been studied in the general regression context.

We begin by describing the method (which we coin Deep BLR), introducing a novel variation which enables learning heteroscedastic noise, thus addressing a weakness of previous applications of BLR to deep representations. The method, along with ensembles of it, is then evaluated on a set of standard datasets, which is common in the literature for BNNs but has not been done for BLR. Finally, we evaluate Deep BLR as an uncertainty-aware model in a model-based RL algorithm, in which it replaces Deep Ensembles. 

\subsection*{Related work}
In prior work \citep{Snoek2015ScalableNetworks,Riquelme2018DeepSampling,Azizzadenesheli2018EfficientQ-networks}, the predictive variance is either assumed known or unknown but homoscedastic. We propose a simple modification which permits learning flexible and heteroscedastic aleatoric uncertainty estimates by incorporating the underlying NN's  variance prediction.

\section{Method}

Given some data $\mathbf{X}\in\mathbb{R}^{N\times p}$ with corresponding target values $\mathbf{y}\in\mathbb{R}^N$ (for multi-output problems, we can, and do, treat the dimensions independently), a NN with a Gaussian output layer (i.e., the NN predicts mean $\mu(\mathbf{x})$ and variance $\sigma^2(\mathbf{x})$ for each input $\mathbf{x}$) is trained by maximizing the log-likelihood. The NN is then used in two ways in the BLR:
\begin{enumerate}
    \item Regression is done on the latent representation $\mathbf{z}(\mathbf{x})$ obtained from the last hidden layer.
    \item The predicted variance $\sigma^2(\mathbf{x})$ is used as a ``known'' variance in the likelihood.
\end{enumerate}

A helpful distinction is between \emph{aleatoric} and \emph{epistemic} unertainty, where aleatoric uncertainty is irreducible uncertainty, e.g., due to noise, and epistemic uncertainty is uncertainty that is due to our lack of knowledge about the world. In this sense, the Gaussian output estimates the aleatoric uncertainty whereas the BLR keeps track of epistemic uncertainty, i.e., our uncertainty in $\mathbf{w}$.

\begin{figure}
\centering
\includegraphics[width=0.35\textwidth]{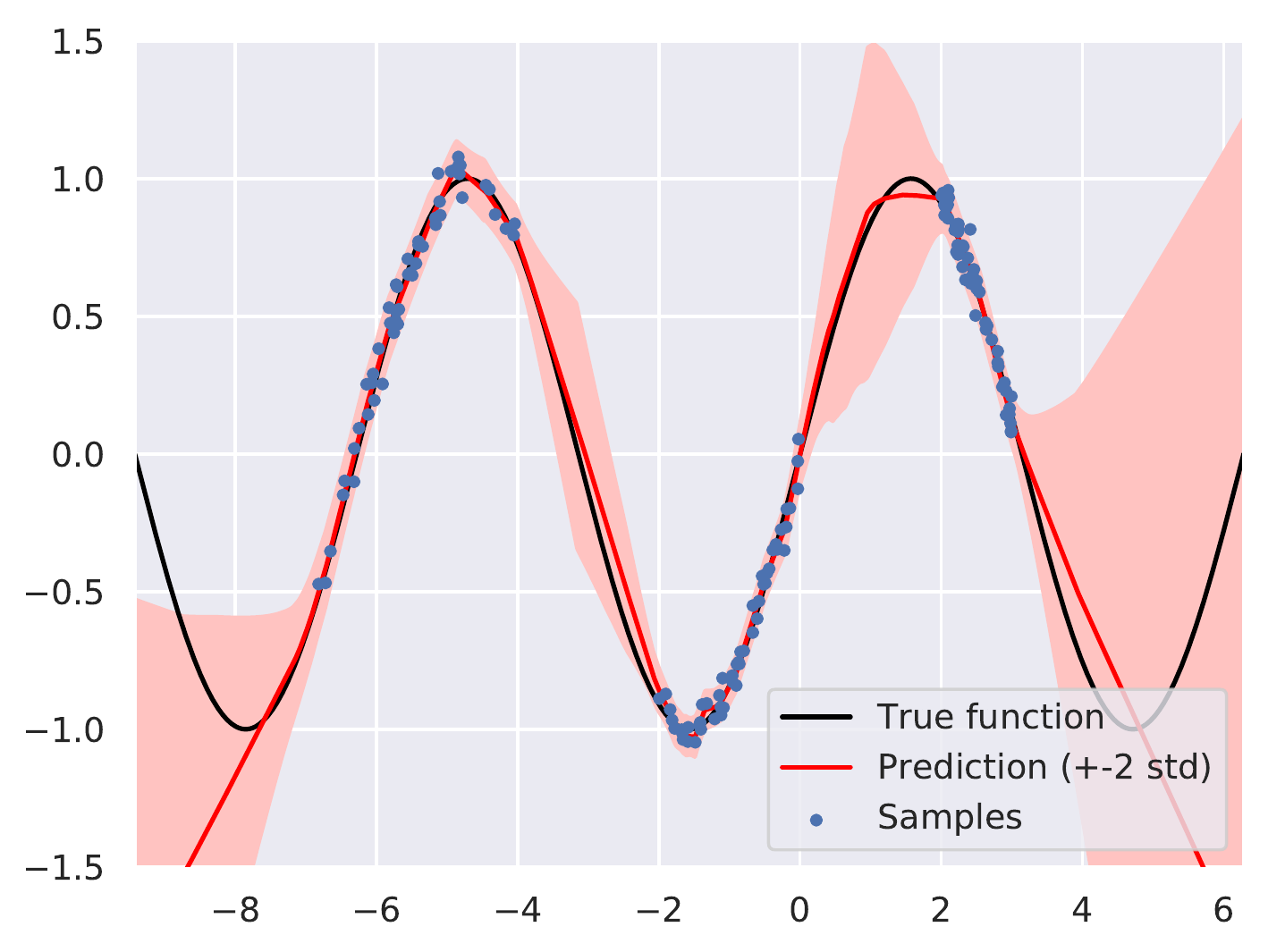}
\caption{The posterior predictive distribution of Deep BLR in a simple toy problem.} \label{fig:blr_example}
\end{figure}

Concretely, we choose a normal prior $p(\mathbf{w})=\mathcal{N}(\mathbf{w}\,|\,\mathbf{0}, g\mathbf{I})$ and a normal likelihood $p(y \,|\, \mathbf{z}, \mathbf{w}, \sigma^2(\mathbf{x})) = \mathcal{N}(y\,|\,\mathbf{z}^T \mathbf{w}, \sigma^2(\mathbf{x}))$, where $g>0$ is a hyperparameter that controls the prior uncertainty. We assume that both inputs and targets are centered for simplicity. Then, by Bayes' rule \citep{Murphy2012MachinePerspective}, the posterior is $\mathcal{N}(\mathbf{w}\,|\,\mathbf{w}_N, \mathbf{V}_N)$ with parameters
\[
\begin{aligned}
\mathbf{V}_N &= \left(\frac1g \mathbf{I} + \mathbf{Z}^T \bm{\Sigma}^{-1}\mathbf{Z}\right)^{-1},\\
\mathbf{w}_N &= \mathbf{V}_N \mathbf{Z}^T \bm{\Sigma}^{-1} \mathbf{y},
\end{aligned}
\]
where $\bm{\Sigma} = \text{diag}(\sigma^2(\mathbf{x}_1),\dots,\sigma^2(\mathbf{x}_N))$ and $\mathbf{Z}=\begin{bmatrix}\mathbf{z}(\mathbf{x}_1),\:\dots\:,\:\mathbf{z}(\mathbf{x}_N) \end{bmatrix}^T$. Furthermore, the posterior predictive distribution at a test point $\mathbf{x}$, with latent representation $\mathbf{z}=\mathbf{z}(\mathbf{x})$, is given by
\[
p(y\,|\,\mathbf{x}, \mathbf{z}, \mathbf{X}, \mathbf{y}) = \mathcal{N}(y\,|\,\mathbf{z}^T\mathbf{w}, \sigma^2(\mathbf{x}) + \mathbf{z}^T \mathbf{V}_N \mathbf{z}).
\]
Here, $\sigma^2(\mathbf{x})$ and $\mathbf{z}^T \mathbf{V}_N \mathbf{z}$ represent the heteroscedastic aleatoric uncertainty and epistemic uncertainty, respectively. Figure \ref{fig:blr_example} contains an example of Deep BLR applied to a toy 1-D regression problem. We see that the predictive distribution looks sensible and that the necessary nonlinear prediction is captured well by the linear regression on the deep representations.

A straightforward extension of Deep BLR is to train an ensemble of $M$ NNs with different random initializations and apply BLR to each resulting representation. The predictive distribution will then be a mixture of Gaussians.

\section{Experiments}
Deep BLR is compared to two prominent alternative methods for uncertainty estimation: MC-dropout \citep{Gal2016DropoutLearning} and Deep Ensembles \citep{Lakshminarayanan2017SimpleEnsembles}. Additionally, we compare to Deep BLR ensembles. We conduct two experiments: regression on standard datasets and application in uncertainty-aware model-based reinforcement learning.

\subsection{Predictive performance on standard datasets}
We evaluate the predictive performance of Deep BLR and Deep BLR ensembles on a set of standard datasets \citep{Dua2017UCIRepository} with the same experimental setup as in prior work on uncertainty-aware NNs \citep{Gal2016DropoutLearning,Lakshminarayanan2017SimpleEnsembles}. See appendix \ref{app:details} for full details. The prior variance $g$ is chosen with a grid search to minimize the negative log-likelihood (NLL) on a validation set, which is cheap since the NN can be reused. 

The resulting performance in terms of NLL, which acts as a proxy measure for the capability to estimate uncertainty, can be seen in Table \ref{table:uci_results}. We see that Deep BLR performs competitively, and Deep BLR ensemble achieves the best performance on all datasets except the last, on which only a single split was done. 

\begin{table}
\caption{Comparison of MC-dropout, Deep Ensembles, Deep BLR, and Deep BLR ensemble on a set of standard datasets. We report the mean and standard error of predictive NLL across 20 random training-test splits (5 for Protein Structure, 1 for Year Prediction MSD). The best method for each dataset is marked in bold.}
\label{table:uci_results}
\centering
\begin{tabular}{lcccc}
\multicolumn{5}{c}{Predictive NLL} \\\midrule
\textbf{Dataset} &\textbf{MC-dropout} & \textbf{Deep Ensembles}  & \textbf{BLR} & \textbf{BLR Ensemble} \\\midrule
Boston Housing & $\mathbf{2.46\pm0.25}$ & $\mathbf{2.41\pm0.25}$ & $\mathbf{2.36 \pm 0.04}$ & $\mathbf{2.37\pm0.05}$ \\
Concrete Strength & $3.04\pm0.09$ & $3.06\pm 0.18$ & $3.01\pm0.04$ & $\mathbf{2.91\pm0.02}$  \\
Energy Efficiency & $1.99\pm0.09$ & $1.38\pm0.22$ & $\mathbf{1.32\pm 0.03}$ & $\mathbf{1.27\pm0.02}$\\
Kin8nm  & $-0.95\pm0.03$ & $-1.20\pm0.02$ & $-1.20\pm0.01$ & $\mathbf{-1.25\pm 0.00}$ \\
Naval Propulsion & $-3.80\pm0.05$ & $\mathbf{-5.63\pm 0.05}$ & $\mathbf{-5.58\pm0.05}$ & $\mathbf{-5.59\pm0.02}$\\ 
Power Plant & $\mathbf{2.80\pm0.05}$ & $\mathbf{2.79\pm 0.04}$ & $\mathbf{2.81\pm0.01}$ & $\mathbf{2.79\pm0.01}$ \\
Protein Structure & $2.89\pm 0.01$ & $2.83\pm 0.02$ & $2.81\pm0.01$ & $\mathbf{2.75\pm 0.01}$\\
Wine Quality & $\mathbf{0.93\pm0.06}$ & $\mathbf{0.94\pm0.12}$ & $1.00\pm0.03$ & $\mathbf{0.90\pm0.02}$\\
Yacht Hydrodynamics & $1.55\pm 0.12$ & $1.18\pm 0.21$ & $\mathbf{0.95\pm0.01}$ & $\mathbf{0.90\pm0.04}$\\
Year Prediction MSD & $3.59\pm\text{NA}$ & $\mathbf{3.35\pm \textbf{NA}}$ & $3.48\pm\text{NA}$ & $3.39\pm\text{NA}$
\end{tabular}
\end{table}

\subsection{Uncertainty-aware model-based reinforcement learning}
PETS \citep{Chua2018DeepModels} is a model-based reinforcement learning algorithm which utilizes an uncertainty-aware dynamics model in the form of an ensemble of NNs. We study the downstream performance of PETS when the ensemble is replaced with Deep BLR, as well as with a Deep BLR ensemble. For simplicity, we use hyperparameters identical to \cite{Chua2018DeepModels}, i.e., the NNs are not tuned for Deep BLR. The prior variance is set to $g=0.1$. We found that larger $g$ values led to unstable learning, so the regularizing effect appears important.

\begin{figure}
\captionsetup[subfigure]{justification=centering}
\centering
\setlength{\lineskip}{\medskipamount}
\subcaptionbox{CartPole.\label{fig:1a}}{\includegraphics[width=0.49\textwidth]{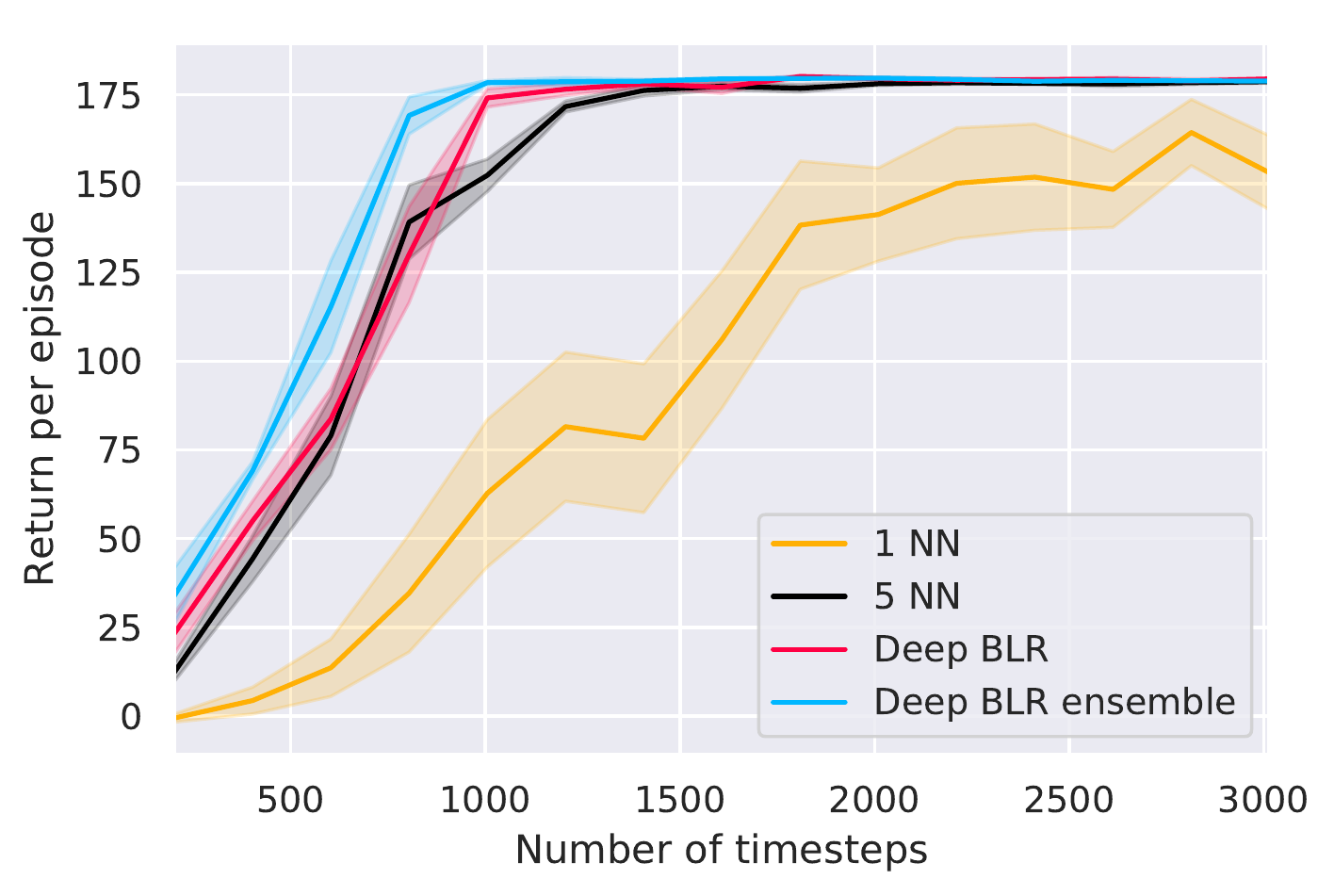}}\hfill
\subcaptionbox{7-dof Reacher.}{\includegraphics[width=0.49\textwidth]{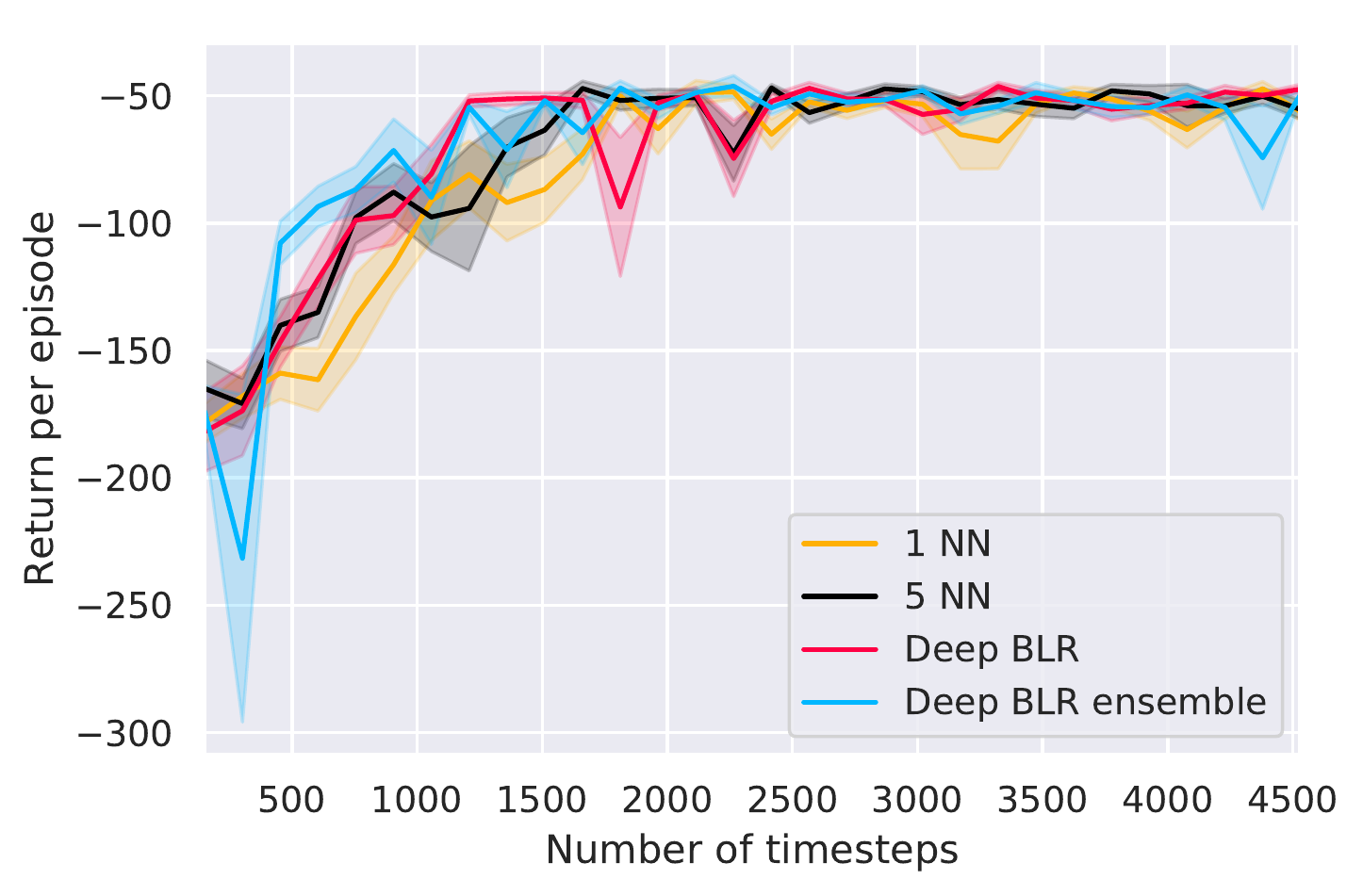}}\hfill
\caption{Downstream comparison of 1 NN, an ensemble of 5 NNs, Deep BLR, and an ensemble of 5 Deep BLR when used as an uncertainty-aware model for PETS on CartPole and 7-dof Reacher in the MuJoCo \citep{Todorov2012MuJoCo:Control} simulator. All NNs parameterize a Gaussian distribution. The experiment is repeated 10 times and the mean return per episode is reported, with the shaded area representing one standard error.} \label{fig:cartpole_results}
\end{figure}

Figure \ref{fig:cartpole_results} shows the results in the CartPole and 7-dof Reacher environment for four different types of models. On CartPole, the poor performance of 1 NN indicates that utilizing epistemic uncertainty is important for efficient learning in that environment. We see that Deep BLR is competitive with, or slightly better than, ensembles of NNs, and that Deep BLR ensemble is the clear winner. On 7-dof Reacher, all methods perform very similarly, indicating that accurate uncertainty estimation does not aid PETS in learning a good policy in this environment. Thus, even if Deep BLR was significantly better than Deep Ensembles, we could not distinguish them in this environment.

\section{Conclusions and future work}
Bayesian linear regression on deep representations is a simple, flexible method for obtaining uncertainty-aware NNs for regression. Our experiments indicate that Deep BLR is competitive with (and as an ensemble can outperform) the commonly used ensemble methods, which is consistent with prior work \citep{Riquelme2018DeepSampling}. Note that BLR was used on top of the existing architectures with no tuning, and further tuning may be beneficial. 

We believe there is much potential for future work in this direction. In particular, investigating the connection between the NN architecture and reliability of BLR uncertainty estimates, and extending the idea to classification. In the case of, e.g., Bayesian logistic regression, there is no analytic posterior, but perhaps approximations suffice. Further studying the downstream performance of Deep BLR would also be interesting, e.g., in uncertainty-aware model-based RL.

\subsubsection*{Acknowledgments}
This work was partially supported by the Wallenberg Al, Autonomous Systems and Software Program (WASP) funded by the Knut and Alice Wallenberg Foundation.

\small

\bibliographystyle{plainnat}
\bibliography{references}

\newpage
\normalsize
\appendix
\section{Experimental details}\label{app:details}
We replicate the experimental setup in prior work \citep{Gal2016DropoutLearning,Lakshminarayanan2017SimpleEnsembles}. All methods use a NN with a single hidden layer with 50 ReLU, except for the larger Protein Structure and Year Prediction MSD datasets where 100 ReLU are used. Each NN parameterizes a normal distribution and is trained to minimize the negative log-likelihood using Adam \citep{Kingma2015Adam:Optimization} for 40 epochs with batch size 32 and learning rate 0.01 (except 0.001 and 0.0001 for Protein Structure and Year Prediction MSD respectively.) All ensembles consist of 5 identical NNs with different random initializations. We evaluate the methods on 20 random 90/10 training/test splits, except for the larger Protein Structure and Year Prediction MSD where only 5 and 1 splits are done respectively. The inputs and outputs are always normalized to have zero mean and unit variance.

\end{document}